\documentclass[letterpaper, 10 pt, conference]{ieeeconf}

\usepackage{mathtools}
\usepackage[pdftex,final]{graphicx}
\usepackage[english]{babel}
\usepackage[latin1]{inputenc}
\usepackage{balance}

\usepackage{url}
\usepackage{amsmath,amsfonts,amssymb}
\usepackage{xcolor}

\usepackage{caption}
\usepackage{subcaption}

\usepackage{algorithm}
\usepackage[noend]{algpseudocode}

\usepackage{textpos}

\usepackage{pbox}
\usepackage{multirow}

\usepackage{setspace}
\let\Algorithm\algorithm
\renewcommand\algorithm[1][]{\Algorithm[#1]\setstretch{1.4}}

\makeatletter
\def\BState{\State\hskip-\ALG@thistlm}
\makeatother

\definecolor{darkblueish}{RGB}{64, 87, 116}
\definecolor{blueish}{RGB}{103, 135, 176}
\definecolor{greenish}{RGB}{177, 177, 123}
\definecolor{reddish}{RGB}{ 205, 102, 7}
\definecolor{orangeish}{RGB}{246, 160, 61}
\definecolor{somethingilike}{RGB}{250, 232, 184}
\colorlet{bluegreenish}{blueish!50!greenish}

\usepackage[bookmarks=true]{hyperref}
\hypersetup{
    colorlinks,
    linkcolor={reddish!70!black},
    citecolor={blueish!70!black},
    urlcolor={blueish!70!black}
}
\usepackage[all]{hypcap}

\usepackage[disable]{todonotes}

\newcommand{\cristina}[1]{\todo[inline,color=red!40]{Cristina: #1}}

\renewcommand{\baselinestretch}{0.976}

\makeatletter
\def\input@path{{../figures/}}
\makeatother
\graphicspath{{../figures/}}

\makeatletter
\let\NAT@parse\undefined
\makeatother
\usepackage[numbers]{natbib}

\IEEEoverridecommandlockouts                             
\newtheorem{exmp}{Example}[section]

\title{\LARGE \bf
Robust Gaussian Filtering using a Pseudo Measurement
}

\author{Manuel W\"uthrich$^{1}$, Cristina Garcia Cifuentes$^{1}$, Sebastian Trimpe$^{1}$\\ 
Franziska Meier$^{1,2}$, Jeannette Bohg$^{1}$, Jan Issac$^{1}$ and Stefan Schaal$^{1,2}$ 
  \thanks{$^{1}$ Autonomous Motion Department at the Max Planck Institute for Intelligent Systems, T\"ubingen, Germany. Email: {\tt\small first.lastname@tuebingen.mpg.de}}%
  \thanks{$^{2}$ Computational Learning and Motor Control lab at the
  University of Southern California, Los Angeles, CA, USA.}%
}

\begin{document}

\maketitle
\thispagestyle{empty}
\pagestyle{empty}

\begin{abstract}
Many sensors, such as range, sonar, radar, GPS and visual devices,
produce measurements which are contaminated by outliers. This
problem can be addressed by using fat-tailed sensor models, which
account for the possibility of outliers.
Unfortunately, all estimation algorithms
  belonging to the family of Gaussian filters (such as the widely-used
  extended Kalman filter and unscented Kalman filter) are inherently
  incompatible with such fat-tailed sensor models.  The contribution
  of this paper is to show that any Gaussian filter can be made
  compatible with fat-tailed 
sensor models by applying one simple change: Instead of filtering with 
the physical measurement, we propose to filter
with a pseudo measurement obtained by applying a feature function
to the physical measurement. 
We derive such a feature function which is
optimal under some conditions. 
Simulation results show that the proposed method can effectively handle 
measurement outliers and allows for robust filtering in both linear and nonlinear systems.
%
%
%
%
%
%
%
%
%
\end{abstract}

\begin{textblock*}{100mm}(.\textwidth,-10.5cm)
 \begin{spacing}{.8}
 {\fontsize{8pt}{2pt}\selectfont \sffamily
\noindent 2016 American Control Conference\\
July 6-8, 2016, Boston, MA, USA}
\end{spacing}
\end{textblock*}%

%

\section{Introduction}\label{sect:introduction}
Robust and accurate state estimation is essential to safely control 
any dynamical system. 
However, many sensors, such as range, sonar, radar, GPS and visual devices,
provide measurements populated with outliers.
Therefore, the estimation algorithm must
not be unduly affected by such outliers.

In this paper we argue that problems with outliers are a direct consequence
of unrealistic, thin-tailed sensor models. Unfortunately, many widely-used 
estimation algorithms are inherently incompatible with more realistic, 
fat-tailed sensor models. This holds true for the extended Kalman filter (EKF) \cite{ekf},
the unscented Kalman filter (UKF) \cite{ukf}, and any other member of the family of Gaussian filters (GF) \cite{sarkka}, as we will show in Section~\ref{sec:gf_fat_tails}.

%
%
The contribution of this paper is to show that any member of the family 
of GFs can be made compatible with fat-tailed sensor 
models by applying one simple change: Instead of filtering with the physical measurement,
we filter with a pseudo measurement. This pseudo measurement is
obtained by applying a time-varying feature function to the physical
measurement. We derive a feature function which is optimal under some 
conditions. In simulation experiments, 
we demonstrate the robustness and accuracy of the proposed method 
for linear as well as nonlinear systems.

Numerous robustification methods have been proposed for individual members
of the family of GFs, often involving significant algorithmic changes.
In contrast, the proposed method can be applied to any GF with only minor
changes in the implementation. 
Any existing GF implementation can be 
robustified by merely replacing the sensor model with a pseudo sensor model,
and the physical measurement with a pseudo measurement.

%
%

\section{Related Work}

Ad-hoc procedures for reducing the influence of outliers
have been employed by engineers for a long time. One such
heuristic is to simply discard all measurements which 
are too far away from the expected measurement. This 
approach lacks a firm theoretical basis and there 
is no rigorous way of choosing the thresholds.
Furthermore, the information contained in measurements 
outside of the thresholds is discarded completely, 
which can lead to decreased efficiency \cite{schick}.
For these reasons, significant research effort
has been devoted to robustifying GFs in a principled 
manner. 
In the following we distinguish two main currents on robust
filtering, the first is based on robust statistics in the sense of \cite{huber1964} 
and the second is based on fat-tailed sensor models. 

\subsection{Robust Statistics}\label{sec:robust_estimators}
In the framework of robust statistics in the spirit of \cite{huber1964}, 
the objective is to find an estimator 
with a small variance when the Gaussian noise is contaminated
with noise from a broad class of distributions. The 
resulting estimators are intermediary between the sample 
mean and the sample median.
For instance, \citet{masreliez} propose such an estimator for linear
systems. This approach is extended by \citet{schick}. 
\cristina{remove: However,
these approaches are restricted to linear systems.}

\subsection{Fat-tailed Sensor Models}
Since fat-tailed sensor models are by definition non-Gaussian,
finding the posterior estimate is not trivial. In particular, 
a lot of effort has been devoted to finding filtering recursions 
for models with Student $t$-distributed noise.

\citet{roth} show that for linear systems where the noise and the state are 
jointly $t$-distributed, an exact filter can be found. The authors mention 
that these noise conditions are rarely met in practice, and 
propose an approximation for state-independent $t$-distributed noise. A 
different approximation scheme for linear systems with $t$-distributed 
noise is proposed in \citet{meinhold}.

While those approximations are hand-crafted, \citet{ting} and 
\citet{sarkka_noise} use variational inference techniques to find an optimal 
approximation to the posterior. \citet{agamennonizzle, agamennoni} unify and generalize 
those methods. 

\subsection{Extensions to Nonlinear Systems}
All methods mentioned above assume a linear sensor model.
It is possible to apply them to nonlinear systems by
linearizing the sensor model at each time step, as is done in the EKF.
However, the EKF has been shown to yield poor performance 
for many nonlinear systems \cite{spkf, ukf, gf}.

Application of these robustification methods to other members
of the family of GFs, such as the UKF or the divided difference filter (DDF) \cite{ddf},
is not straightforward. 

One way of doing so is proposed by \citet{karlgaard}, who use a robust
Huber estimator \cite{huber1964} in a DDF. 
Similarly, \citet{piche} propose a method of extending the 
mentioned linear Student $t$-based filtering methods to nonlinear GFs.
However, both of these methods rely on an iterative 
optimization at each time step, which is computationally expensive.
In contrast, the robustification proposed in this paper
allows to robustify any of the numerous GF algorithms with just
minor changes in the implementation.

\section{Filtering}
A discrete-time state-space model 
can be defined by two 
probability distributions: a transition model 
$p(x_t|x_{t-1})$,
which describes the evolution of the state in time, 
and a sensor model 
$p(y_t|x_t)$,
which describes how the measurement 
$y_t$ is generated given the state $x_t$. Alternatively, these 
two models can also be written in functional form
\begin{align}
 x_t &= g(x_{t-1},v_t)\label{eq:transition_functional}\\
 y_t &= h(x_t, w_t)\label{eq:measurement_functional}
\end{align}
with $v_t$ and $w_t$ being 
normally distributed noise variables.
Note that any (even non-Gaussian) model can be specified in this way,
since $v_t$ and $w_t$ can be mapped onto any desired distribution inside
the nonlinear functions $g(\cdot)$ and $h(\cdot)$.

\subsection{Exact Filtering}
Filtering is concerned with estimating the current state $x_t$ 
given all past measurements $y_{1:t} = \{y_1, \dots, y_t\}$. 
The posterior distribution of the current state $p(x_{ t }|y_{ 1:t })$ 
can be computed recursively from the distribution of the previous state 
$p(x_{ t-1 }|y_{ 1:t-1})$. This recursion can be written in two steps: 
a prediction step\footnote{We use the notation $\int_x(\cdot)$ as an abbreviation for 
$\int_{-\infty}^\infty(\cdot) \, dx$.} 
\begin{align}
 p(x_{ t }|y_{ 1:t-1 })=
 \int_{ x_{ t-1} } p(x_{ t }|x_{ t-1})p(x_{ t-1 }|y_{ 1:t-1 })
 \label{eq:prediction}
\end{align}
and an update step
\begin{align}
 p(x_{ t }|y_{ 1:t })=\frac { p(y_{ t }|x_{ t })p(x_{ t }|y_{ 1:t-1 }) }
 { \int_{ x_{ t } } p(y_{ t }|x_{ t })p(x_{ t }|y_{ 1:t-1 }) } .\label{eq:update}
\end{align}
These equations can generally not be solved in closed form \cite{earlyKushner}. 
The most notable exception is the Kalman filter (KF) \cite{kalman1960new}, 
which provides the exact solution for linear Gaussian systems. 
Significant research effort has been invested into generalizing the 
KF to nonlinear dynamical systems. 

\subsection{Gaussian Filtering}\label{sec:gf}
The KF and its generalizations
to nonlinear systems (e.g. the EKF and the UKF) 
are members of the family of GFs \cite{sarkka, new_perspective, gf, wu}.
GFs approximate both the predicted belief \eqref{eq:prediction},
as well as the posterior belief \eqref{eq:update} with Gaussian distributions.

In the prediction step \eqref{eq:prediction}, the exact distribution
is approximated by a Gaussian\footnote{$\mathcal{N}(z|\mu,\Sigma)$ denotes the Gaussian
with mean $\mu$ and covariance $\Sigma$.}
\begin{align}
p(x_{ t }|y_{1:t-1})=\mathcal{N}(x_t|\mu_{x_t},\Sigma_{x_tx_t}). \label{eq:approx_prediction}
\end{align}
The prediction step is not affected by 
the type of sensor model used and will therefore not be discussed 
here, see for instance \cite{sarkka, new_perspective, gf, wu} 
for more details.

We will only consider the update step \eqref{eq:update} in 
the remainder of the paper. For ease of notation, we will not write 
the dependence on past measurements $y_{1:t-1}$ explicitly anymore.
The remaining variables all have time index $t$, which can therefore
be dropped. The predicted belief $p(x_{ t }|y_{ 1:t-1 })$ 
can now simply 
be written as $p(x)$, and the posterior belief $p(x_{ t }|y_{ 1:t })$ as $p(x|y)$, 
etc.

As shown in \cite{new_perspective}, the GF 
can be understood as finding an approximate Gaussian posterior $q(x|y)$ 
by minimizing the Kullback-Leibler divergence \cite{mckay}
to the exact joint distribution
\begin{align}\label{eq:objective}
 \arg \min _q \textrm{KL}[p(x,y)|q(x|y)].
\end{align}
 The form of 
$q(x|y)$ is restricted to be Gaussian in $x$ 
\begin{align}\label{eq:gf_form}
 q(x|y)=\mathcal{N}\left(x|m(y) ,\Sigma \right)
\end{align}
with the mean being an 
affine function of $y$
\begin{align}
 m(y) = M \begin{pmatrix} 1 \\ y \end{pmatrix}.\label{eq:linear_mean_fct}
\end{align}

This minimization is performed at each update step and yields  
the optimal parameters of the approximation \eqref{eq:gf_form}
\begin{align}
M &=\begin{pmatrix} \mu _{ x }-\Sigma _{ xy }\Sigma _{ yy }^{ -1 }\mu _{ y} & \Sigma _{ xy }\Sigma _{ yy }^{ -1 }  \end{pmatrix}
\\
\quad \Sigma&=\Sigma _{ xx }-\Sigma _{ xy}\Sigma _{ yy }^{ -1 }\Sigma _{ xy }^\intercal  . \label{eq:gamma}
\end{align}
See \cite{new_perspective} for a detailed derivation of this result. 
The parameters $\mu_x$ and $\Sigma_{xx}$ are given by the belief 
\eqref{eq:approx_prediction} computed in the prediction step. 
The remaining parameters are defined as
   \begin{align}
\mu _{ y }&=\int_{ y } y p(y)\label{eq:muy}\\ 
\Sigma _{ yy }&=\int_{ y } (y-\mu _{ y })(y-\mu _{ y })^\intercal p(y)\label{eq:sigmay}\\ 
\Sigma _{ xy }&=\int_{ x,y } (x-\mu _{ x })(y-\mu _{ y })^\intercal p(x,y).\label{eq:sigmaxy}
\end{align}
For a linear system, this solution corresponds to the KF
equations \cite{new_perspective}.

\subsubsection*{Numeric Integration Methods}\label{sec:numeric_integration}
For most nonlinear systems, the integrals \eqref{eq:muy}, \eqref{eq:sigmay} 
and \eqref{eq:sigmaxy} cannot be computed in closed form and have to be approximated.
In the EKF, this is done by linearization
 at the current mean
estimate of the state $\mu_x$. This approximation does not take 
the uncertainty in the estimate into account, which can lead to large errors 
and sometimes even divergence of the filter \cite{spkf, gf}.

Therefore, approximations based on numeric integration methods
are preferable in most cases. Deterministic Gaussian integration
schemes have been investigated thoroughly, and the resulting
filters are collected under the term Sigma Point Kalman Filters 
(SPKF) \cite{spkf}. Well known members of this family are the 
UKF \cite{ukf}, the DDF \cite{ddf} 
and the cubature Kalman filter (CKF) \cite{ckf}. Alternatively,
numeric integration can also be performed using Monte Carlo methods.
The method presented in this paper applies to any GF, regardless of which
particular integration method is used.

\section{A Case for Fat Tails}
Measurement acquisition is typically modeled by a Gaussian or some
other thin-tailed sensor model. This assumption is usually made for
analytical convenience, not because it is an accurate representation
of the belief of the engineer. If an engineer were to believe that
measurements are in fact generated by a Gaussian distribution, then
she would have to accept a betting ratio of $7\times 10^{14}$ to $1$
that no measurement further than $8$ standard deviations from the
state will occur.\footnote{According to De Finetti's definition of
  probability \cite{definetti}.}  Few engineers would be interested in
such a bet, since one can usually not exclude the possibility of
acquiring a large measurement due to unexpected physical effects in
the measurement process.

The mismatch between the actual belief and the Gaussian model can lead to 
counter-intuitive behavior of the inference algorithm. More concretely,
the posterior mean is an affine function of the measurement. This 
implies that the shift in the mean produced by a single measurement is not bounded.

This problematic behavior disappears when using a more realistic, 
fat-tailed model instead of the Gaussian model \cite{meinhold}. There are several definitions of 
fat-tails which are commonly used \cite{fat_tails}. Here, we simply 
mean any distribution which decays slower than the Gaussian. Which 
particular tail model is used depends on the application.

\subsection{The Gaussian Filter using Fat Tails}\label{sec:gf_fat_tails}
The GF approximates all beliefs with Gaussians, but the sensor model can have 
any form. In principle, nothing prevents us from using the GF with a 
fat-tailed sensor model. Unfortunately, the GF is not able to do proper 
inference using such a model. The sensor model $p(y|x)$ enters the GF equations 
only through \eqref{eq:muy}, \eqref{eq:sigmay} and \eqref{eq:sigmaxy}. 
To make this dependency explicit, we substitute $p(y) = \int_x p(y|x)p(x)$ in 
\eqref{eq:muy} and \eqref{eq:sigmay}, and $p(x,y) = p(y|x)p(x)$ in \eqref{eq:sigmaxy}, and integrate in $y$
\begin{align}
\mu _{ y }&=\int _{ x } \mu _{ y|x }(x)p(x)\\ 
\Sigma _{ yy }&=\int _{ x } (\Sigma _{ yy|x }(x)+\mu _{ y|x }(x)\mu _{ y|x }(x)^\intercal -\mu _{ y }\mu _{ y }^\intercal )p(x)\\ 
\Sigma _{ xy }&=\int _{ x } (x-\mu _{ x })(\mu _{ y|x }(x)-\mu _{ y })^\intercal p(x).
\end{align}
What is important to note here is that these equations only depend on the 
sensor model through the conditional mean 
and the conditional covariance
\begin{align}
\mu _{ y|x }(x) &= \int_y y p(y|x) \\
 \Sigma _{ yy|x }(x)&=\int _{ y } (y-\mu _{ y|x }(x))(y-\mu _{ y|x }(x))^\intercal p(y|x).
\end{align}
Since fat-tailed sensor models typically have very large or even 
infinite covariances, the GF will behave as if the measurements were 
extremely noisy. It achieves robustness by simply discarding all 
measurements, which is obviously not the behavior we were hoping for.

\subsection{Simulation Example}
\begin{figure}[tb]
  ~~
  \includegraphics[width=0.85\linewidth]{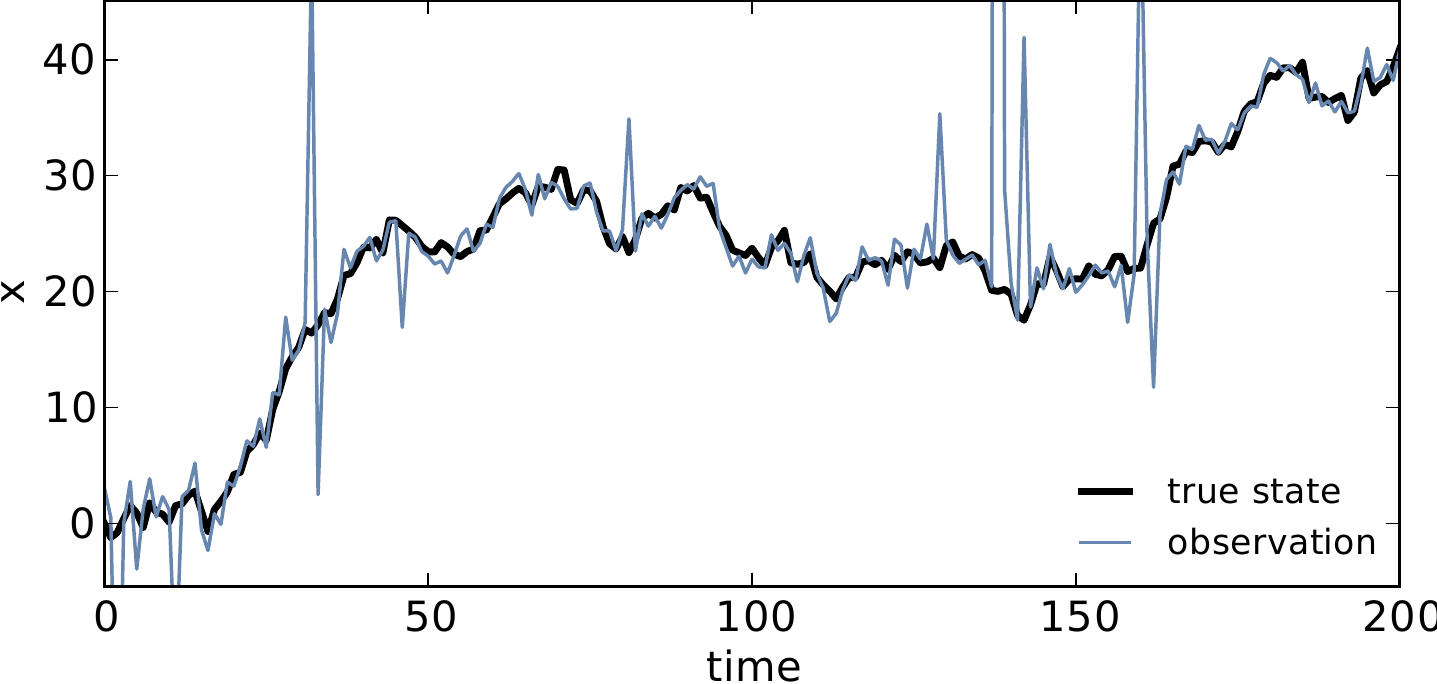}
  \caption{Simulation of the system with fat-tailed measurement described in
    Example~\ref{system_specification}. \label{fig:fat_tails:measurement}}
\end{figure}
\begin{figure}[tb]
  \begin{subfigure}[b]{\linewidth}
    ~~
    \includegraphics[width=0.85\linewidth]{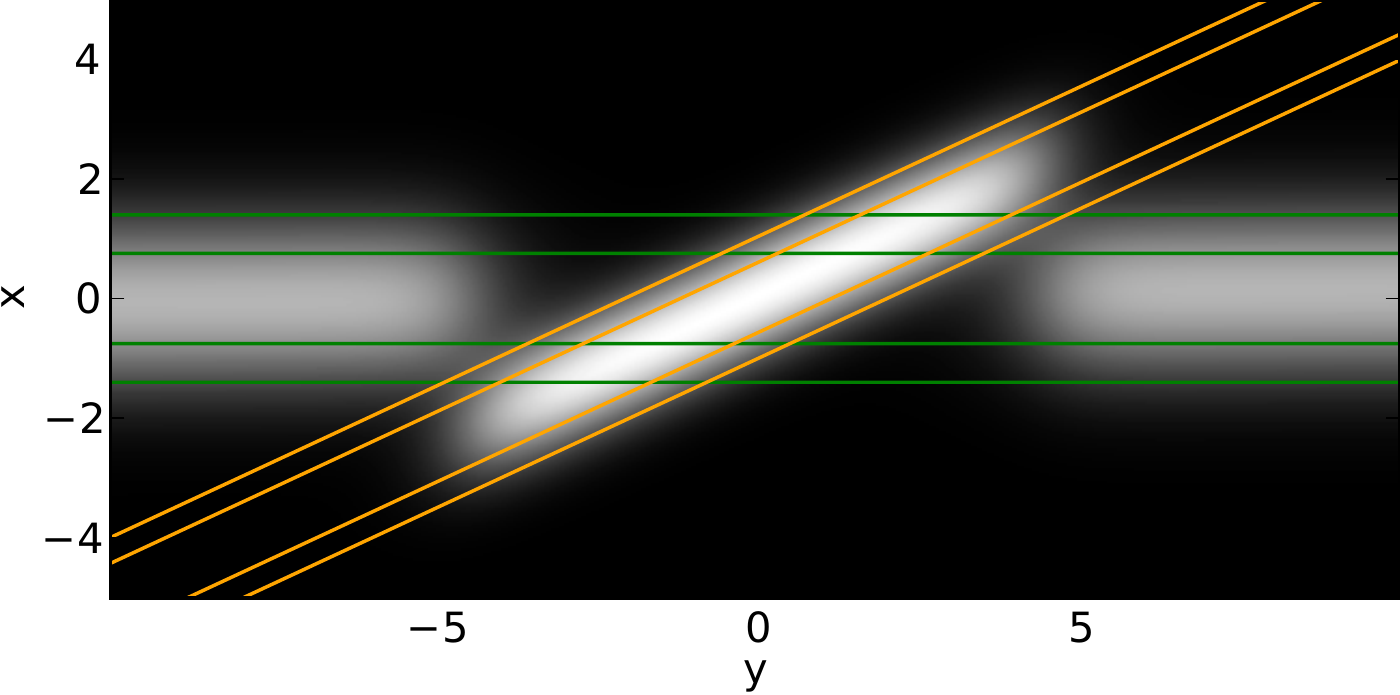}
	\caption{The exact density $p(x_1|y_1)$ (white means higher). Overlaid:
          contour lines of the approximate density $q(x_1|y_1)$ given by a
          fat-tailed GF (green) and a thin-tailed GF
          (orange). \label{fig:gf_fat_tails:density}}
  \end{subfigure}\\
  ~\\
  \begin{subfigure}[b]{\linewidth}
    ~~
    \includegraphics[width=0.87\linewidth]{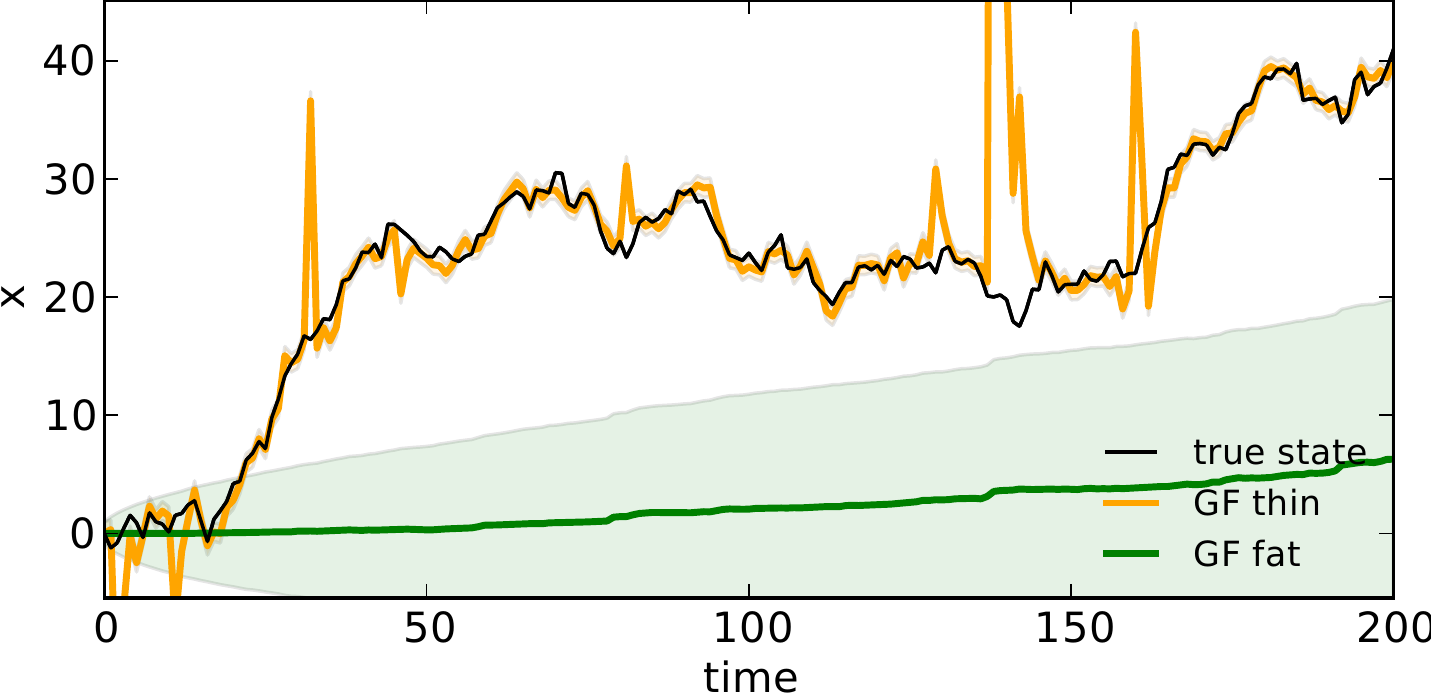}
	\caption{True state of the system over time, with the obtained estimates
          and their standard deviation. \label{fig:gf_fat_tails:time}}
  \end{subfigure}
  \caption{Standard GF applied to the system with fat-tailed measurement
    described in Example~\ref{system_specification}. \label{fig:gf_fat_tails}}
\end{figure}
To illustrate this problematic behavior, we apply the GF to the following
dynamical system:
\begin{exmp}\label{system_specification}
System specification\footnote{$\mathcal{C}(z|\mu,\gamma)$ denotes the Cauchy
  distribution with location $\mu$ and scale $\gamma$.}
  \begin{align}
    p(x_t|x_{t-1})&=\mathcal{N}(x_t|x_{t-1}, 1.0)\\
    p(y_t|x_{t}) &= 0.9 ~ \mathcal{N}(y_t|x_{t}, 1.0)+ 0.1 ~ \mathcal{C}(y_t|x_t,10.0) \label{eq:simulated_sensor}\\
    p(x_0)&=\mathcal{N}(x_0|0.0,1.0)
  \end{align}
\end{exmp}

\vspace{0.3cm}
The measurements are contaminated with Cauchy-distributed noise,
which leads to occasional outliers, as shown in Figure~\ref{fig:fat_tails:measurement}.
We apply two GFs to this problem. The first uses a sensor model
which does not take into account the fat-tailed Cauchy noise, it only
models the Gaussian noise, i.e. the left term in \eqref{eq:simulated_sensor}.
The second GF uses a sensor model
which is identical to the true sensor \eqref{eq:simulated_sensor}.
We will refer to the first filter as the thin-tailed GF, and to the second filter
as the fat-tailed GF.

In Figure~\ref{fig:gf_fat_tails:density},
we show the exact density $p(x_1|y_1)$ after the first filtering step. 
The approximations obtained by the thin-tailed GF (yellow) and 
the fat-tailed GF (green) are overlaid. It can be seen that 
the approximation to the exact posterior is very poor in both cases.
%
%
The mean of the exact density $p(x|y)$ is approximately 
linear in $y$ for small $y$. For measurements $y$ larger
than about $5.0$, the posterior mean reverts back to the prior mean
$0.0$ and does not depend on $y$ anymore.

This behavior cannot be captured by an approximation of 
the form of \eqref{eq:linear_mean_fct}, since it only allows for linear dependences 
in $y$. 
The approximation by the thin-tailed GF fits the exact
posterior well for small $y$, but instead of flattening out it keeps growing linearly
for large $y$. Hence, it is not robust to outliers.
The approximation by the fat-tailed GF correctly
captures the behavior of the exact posterior for large $y$,
i.e. it is independent of $y$. However, this implies that
all measurements, not just outliers, are ignored, 
as expected from the analysis in Section~\ref{sec:gf_fat_tails}.
For both filters, the poor fit translates to poor 
filtering performance, as shown in Figure~\ref{fig:gf_fat_tails:time}.

\section{A Measurement Feature for Robustification}
To enable the GF to work with fat-tailed sensor models,
we hence have to change the form of the approximate belief \eqref{eq:gf_form}.
In \cite{new_perspective} it is shown that more flexible approximations 
can be obtained by allowing for nonlinear features in $y$. The mean function
\eqref{eq:linear_mean_fct} then becomes
\begin{align}\label{eq:feature_mean_fct}
m(y) = M \begin{pmatrix} 1 \\ \varphi(y) \end{pmatrix}.
\end{align}
The resulting filter is equivalent to the standard GF using a virtual 
measurement which is obtained by applying a nonlinear feature function 
$\varphi(\cdot)$ to the physical measurement.
\cristina{removed: \\
$y$. This feature function can be time-varying, and may depend on
the current belief}

In the following, we find a feature $\varphi(\cdot)$ 
which enables the GF to work with fat-tailed sensor models.
Instead of hand-designing such a feature, we attempt to find a feature which
is optimal in the sense that it minimizes the KL divergence between
the exact and the approximate distribution \eqref{eq:objective}.

For this purpose, we first find the optimal, non-parametric mean function
$m^*(y)$ with respect to \eqref{eq:objective}. 
Knowing that the mean $m(y)$ is an affine function 
\eqref{eq:feature_mean_fct} of the feature $\varphi(y)$, we 
can then deduce the optimal feature function $\varphi^*(y)$.

\subsection{The Optimal Mean Function}
 In order to find the function $m^*(y)$ which minimizes \eqref{eq:objective},
 we rewrite the objective \eqref{eq:objective}
\begin{align}
 \textrm{KL}[p(x,y)|q(x|y)]&=\int_{ x,y } \log  \left( \frac { p(x,y) }{ q(x|y) }  \right) p(x,y)\\ 
 &=\int_{ y } \textrm{KL}[p(x|y)|q(x|y)]p(y)+C \label{eq:new_objective}
\end{align}
where we have collected  the terms independent of $q(x|y)$ in $C$. 
Since there is no constraint on $m(y)$, \eqref{eq:new_objective}
can be 
optimized for each $y$ independently. 
This means that the integral can be dropped, and we can simply minimize 
the integrand $\textrm{KL}[p(x|y)|q(x|y)]$ with respect to $m(y)$. 
It is a standard result from variational inference that the optimal
parameters of a Gaussian approximation are obtained by moment matching 
\cite{barber}. That is, the optimal mean function $m^*(y)$ of
the approximation is simply equal to the exact posterior mean
\begin{align}
 m^*(y) = \mu_{x|y}(y) = \int_x x p(x|y).
\end{align}
Therefore, the feature vector $\varphi(y)$ would ideally be chosen such that 
$\mu_{x|y}(y)$ can be expressed through a linear combination of features. 
Unfortunately, $\mu_{x|y}(y)$ cannot be found in closed form in most cases. 

The standard GF represents the mean of the posterior as an affine function 
of $y$. This form is optimal for linear Gaussian systems, and it serves as 
a good approximation for many nonlinear thin-tailed systems. Similarly, the 
idea here is to find the optimal feature for a linear Gaussian system with 
an additive fat tail. This feature can be expected to provide a good approximation 
for nonlinear fat-tailed systems.

\subsection{The Optimal Feature for a Linear, Fat-Tailed Sensor}
Suppose that we have a linear Gaussian sensor model 
\begin{align}
b(y|x)=\mathcal{N}(y|Ax+a,P)
\end{align}
which we refer to as the body. 
We would like to add a fat tail $t(\cdot)$ to make the filter robust to outliers. 
The combined sensor model with tail weight $0\le \omega \le 1$ is then
  \begin{align}
  \boxed{
  p(y|x)=(1-\omega)b(y|x) + \omega t(y|x).\label{eq:combined_observe}
  }
  \end{align}

\subsubsection{Assumptions on the Form of the Tail}
The precise shape of the tail is application specific
and does not matter for the ideas in this paper. However, 
the subsequent derivation relies on the assumption that the 
tail $t(y|x)$ is almost constant in $x$
on the length scale of the standard deviation of the belief $p(x)$.
This allows us to treat $p(x)$ like a Dirac function
with respect to $t(y|x)$.
More concretely, we will assume that the approximation
\begin{align}
 \int_x f(x) t(y|x) p(x) \approx t(y|\mu_x)\int_x f(x) p(x), \label{eq:approx_tail_integral}
\end{align}
is accurate for any affine function $f(x)$.

This is a reasonable assumption, since the tail accounts for unexpected
effects in the measurement process, which by definition bear little or no
relation to the state $x$.
For instance, \citet{thrun_monte_carlo} suggest to use a tail which is independent
of the state $x$ and uniform in $y$, to account for outliers in range sensors.
For such uniform tails, \eqref{eq:approx_tail_integral} is exact. For 
state-dependent tails, we expect this approximation
to be accurate enough to provide insights into 
the required form of the feature. 
  
\subsubsection{The Conditional Mean}
We will now find the posterior mean $\mu_{x|y}(y)$ for this measurement 
model, which will then allow us to find the optimal feature. 
The posterior mean can be obtained from the predicted belief $p(x)$
and the sensor model $p(y|x)$ using Bayes' rule
\begin{align}
 \mu _{ x|y }(y) = \int_x x p(x|y) = \frac{\int_x xp(y|x)p(x)}{\int_x p(y|x)p(x)}.
\end{align}
Inserting \eqref{eq:combined_observe} we obtain
  \begin{align}
\mu _{ x|y }(y)=\frac { (1-\omega )\int _{ x } xb(y|x)p(x)+
\omega \int _{ x } xt(y|x)p(x) }{ (1-\omega )\int _{ x } b(y|x)p(x)+
\omega \int _{ x } t(y|x)p(x) }.
\end{align}
Both the predicted belief $p(x)=\mathcal{N}(x|\mu_x,\Sigma_{xx})$ and the 
body of the sensor model $b(y|x)$ are Gaussian. Therefore, the integrals 
in the first term of the numerator and the first term in the denominator can be 
solved analytically using standard Gaussian marginalization and conditioning. 
The integrals in the second terms of the numerator and the 
denominator can be approximated according to \eqref{eq:approx_tail_integral}, and
we obtain
\begin{align}
\mu &_{ x|y}(y)\approx \tilde{\mu} _{ x|y }(y)\\
&=\frac { (1-\omega )({ d+Dy })
\mathcal{N}({ y }|\mu _{y}^b,\Sigma _{yy}^b)+\omega \mu _{ x }t(y|\mu _{ x }) }
{ (1-\omega )\mathcal{N}({ y }|\mu _{y}^b,\Sigma _{yy}^b)+\omega t(y|\mu _{ x }) } 
\label{eq:optimal_mean}
\end{align}
where we have defined 
\begin{align}
D&=({ \Sigma _{ xx }^{ -1 } }+{ A^\intercal }{ P^{ -1 } }{ A })^{ -1 }
{ A^\intercal }{ P^{ -1 } }\\ 
d&=({ \Sigma _{ xx }^{ -1 } }+{ A^\intercal }{ P^{ -1 } }{ A })^{ -1 }
({ \Sigma _{ xx }^{ -1 } }{ \mu _{ x } }-{ A^\intercal }{ P^{ -1 } }{ a) }.
\end{align}
The expectations in \eqref{eq:optimal_mean}
\begin{equation}
   \boxed{
   \begin{aligned}
\mu _{y}^b&=\int _{ x } \int _{ y } yb(y|x)p(x)\\
\Sigma _{yy}^b&= \int _{ x } \int _{ y } (y-\mu _{y}^b)(y-\mu _{y}^b)^\intercal b(y|x)p(x)
   \end{aligned}
  } \label{eq:body_expectations}
\end{equation}
only involve the body, and not the tail distribution. Hence, we avoid
the problems related to the potentially huge or even infinite covariance of the tail discussed in 
Section~\ref{sec:gf_fat_tails}.

\begin{figure}[tb]
  \centering
  	\includegraphics[width=0.7\linewidth]{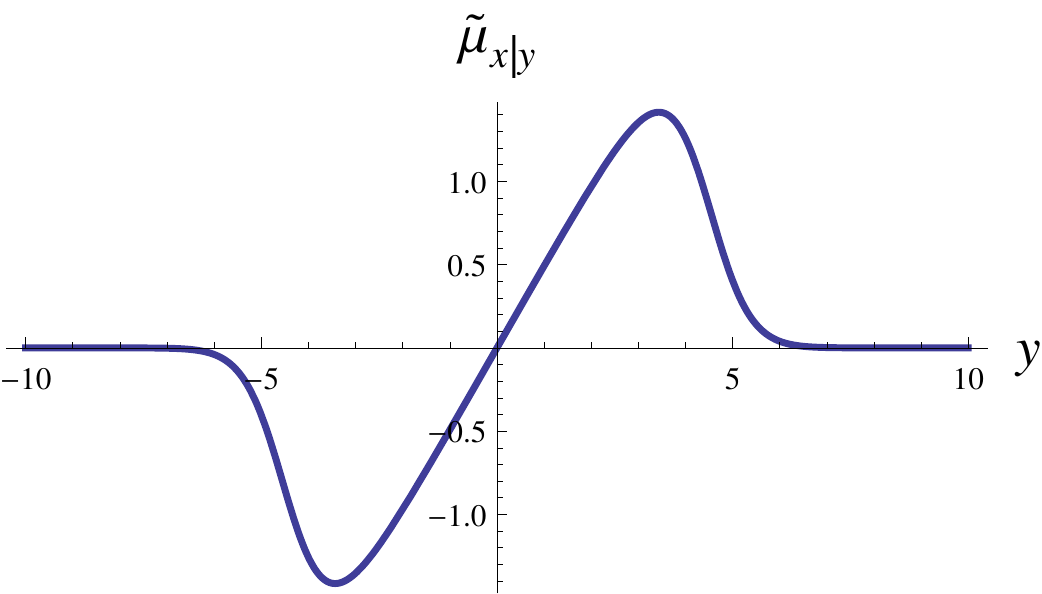}
 	 \caption{The approximate optimal mean function
\eqref{eq:optimal_mean} for the dynamical system
from Example \ref{system_specification} (at time $t=1$).}
  	\label{fig:mean_function}
\end{figure}

In Figure \ref{fig:mean_function}, we plot the optimal mean function
\eqref{eq:optimal_mean} for dynamical system
in Example \ref{system_specification} (at time $t=1$). 
For $y$ close to the expected measurement 
$\mu_y=0$, the conditional mean \eqref{eq:optimal_mean} 
is approximately linear in $y$. If a measurement $y$ of large
magnitude 
is obtained, then the tail becomes predominant, and the posterior mean 
reverts to the prior mean $\mu_x=0$. 

The standard GF attempts to approximate this function by an affine 
function \eqref{eq:linear_mean_fct}. Not surprisingly, this yields
very poor results, as shown in Figure \ref{fig:gf_fat_tails:density}.

\subsubsection{The Optimal Feature}
To identify the feature required to express the optimal
mean $m^*(y)=\mu_{ x|y}(y)\approx \tilde{\mu}_{ x|y}(y)$, we compare \eqref{eq:optimal_mean} to \eqref{eq:feature_mean_fct}.
All the constant terms can be collected in 
$
 M =  \begin{pmatrix} 0 & d & D & \mu _{ x } \end{pmatrix}$
and all the terms which depend on $y$ are part of the feature\footnote{
The factors $(1-\omega)$ and $\omega$ in the numerator could equally well 
have been collected in $M$ instead of the feature, since they are constant.
However, we prefer to maintain these terms in the feature since they
provide appropriate scaling.}
\begin{align}
\boxed{
 \varphi (y)=\frac { 
 \begin{pmatrix} 
 (1-\omega )\mathcal{N}({ y }|\mu _{y}^b,\Sigma _{yy}^b) \\ 
 y(1-\omega )\mathcal{N}({ y }|\mu _{y}^b,\Sigma _{yy}^b) \\ 
 \omega t(y|\mu _{ x }) 
 \end{pmatrix} }
 { (1-\omega )\mathcal{N}({ y }|\mu _{y}^b,\Sigma _{yy}^b)+\omega t(y|\mu _{ x }) }.\label{eq:optimal_feature}
 }
\end{align}

\begin{figure}[tb]
  \centering
  	\includegraphics[width=0.7\linewidth]{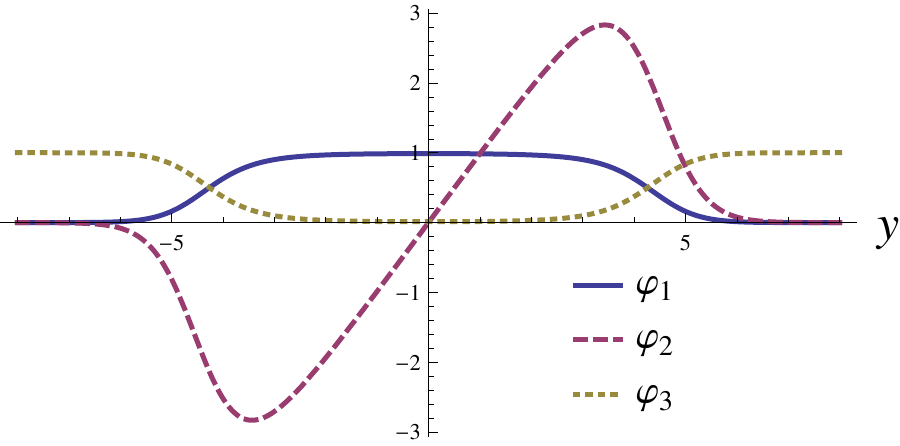}
 	 \caption{The three components of the optimal feature
\eqref{eq:optimal_feature} for the dynamical system
from Example \ref{system_specification} (at time $t=1$).}
  	\label{fig:features}
\end{figure}

%
%
%
In Figure \ref{fig:features}, we plot the three dimension of \eqref{eq:optimal_feature}
for the Example \ref{system_specification}.
All of the feature components are asymptotically constant in $y$, which means
that the estimate remains bounded for arbitrarily large measurements. The 
three components have intuitive interpretations. The first two
components are approximately constant and linear 
 in $y$ respectively, for measurements close to the expected value. 
Hence, they allow the filter 
to express an affine dependence on $y$ which will vanish for very large 
measurements.
The third component is small for $y$ close to the expected value, and grows up
to some constant for $y$ which are large. It hence allows the mean estimate 
to revert to a constant value for large measurements.

For the special case of $\omega=0$, the feature becomes
\begin{align}
 \varphi(y)=(1,y,0)^\intercal.\label{eq:feature_collapse}
\end{align}
Thus, if the sensor model does not have a fat tail, the standard Gaussian Filter
is retrieved. The linear mean function \eqref{eq:linear_mean_fct} is 
a special case of the feature mean function \eqref{eq:feature_mean_fct}.
%

\section{The Robust Gaussian Filter}
In the previous section, we found the approximately optimal measurement feature 
for a linear Gaussian sensor model with additive fat tails.
The GF can hence be enabled to work with fat-tailed sensor models
by filtering in feature space. This robustification can be applied to 
any member of the family of GFs, 
be it the EKF or an SPKF.
We will refer to the filter obtained
by using the feature \eqref{eq:optimal_feature} as the robust Gaussian 
filter (RGF).

For nonlinear, fat-tailed models, the RGF will not be optimal, but it 
provides a good approximation in the same way the standard GF provides a 
good approximation to nonlinear, thin-tailed sensor models. If 
the RGF is applied to a sensor model without a fat tail,
it will coincide with the standard GF, since the feature reduces
to a linear function \eqref{eq:feature_collapse}. Hence,
the RGF extends the GF. It broadens its domain of applicability
to fat-tailed sensor models.

\subsubsection*{Algorithm}
For clarity, we describe the RGF algorithm here step by step. Since
this involves variables of several time steps, we will reintroduce
the time indices which we dropped earlier.

The standard GF is described in Algorithm \ref{alg:gf}.
\begin{algorithm}[!tb]
\caption{Gaussian Filter}
\label{alg:gf}
\begin{algorithmic}[1]
\Require $p(x_{ t - 1 }| y_{1:t-1}),y_t,g(\cdot),h(\cdot)$
\Ensure  $p(x_{ t }| y_{1:t})$
\State $p(x_{ t }| y_{1:t-1}) = \text{\tt predict}[p(x_{ t - 1 }| y_{1:t-1}),g(\cdot)]$
\State $p(x_{ t }| y_{1:t}) = \text{\tt update}[p(x_{ t}| y_{1:t-1}),h(\cdot),y_t]$
\State Return $p(x_{t}|y_{1:t})$
\end{algorithmic}
\end{algorithm}
The input to 
the algorithm are the previous belief, the new measurement $y_t$,
the transition model \eqref{eq:transition_functional} and
the sensor model \eqref{eq:measurement_functional}. The GF simply
predicts, then updates, and finally returns the new estimate.
The concrete implementation of the $\text{\tt predict}$ and 
the $\text{\tt update}$ functions depends on whether we are using
an EKF, a UKF, a DDF or some other GF.

\begin{algorithm}[!tb]
\caption{Robust Gaussian Filter}
\label{alg:rgf}
\begin{algorithmic}[1]
\Require $p(x_{ t - 1 }| y_{1:t-1}),y_t,g(\cdot),h(\cdot),h^b(\cdot),t(\cdot),\omega$
\Ensure  $p(x_{ t }| y_{1:t})$
\State $p(x_{ t }| y_{1:t-1}) = \text{\tt predict}[p(x_{ t - 1 }| y_{1:t-1}),g(\cdot)]$
\State $\mathcal{N}(y_t|\mu_{y_t}^b,\Sigma_{y_t y_t}^b) = \text{\tt predict}[p(x_{ t }| y_{1:t-1}), h^b(\cdot)]$
\State $\varphi_t(\cdot) = \text{\tt feature}[\mu_{x_t}, \mu_{y_t}^b,\Sigma_{y_t y_t}^b, t(\cdot), \omega]$ 
\Comment{\small as in  \eqref{eq:optimal_feature}, given $\mu_{x_t}$ from Step 1 and $\mu_{y_t}^b,\Sigma_{y_t y_t}^b$ from Step 2.}
\State $p(x_{ t }| y_{1:t}) = \text{\tt update}[p(x_{ t}| y_{1:t-1}),\varphi_t(h(\cdot)),\varphi_t(y_t)]$
\State Return $p(x_{t}|y_{1:t})$
\end{algorithmic}
\end{algorithm}
The RGF is described in Algorithm \ref{alg:rgf}. It requires the same inputs as the GF,
and additionally 
the separate components of the sensor model: body, tail, and tail weight. In particular, the
functional form of the body $h^b(\cdot)$ is used in Step 2, while the feature
computation in Step 3 requires the tail
weight $\omega$ and the evaluation of the tail's distribution $t(\cdot)$.

The RGF delegates all the main computations to the basic GF
through the $\text{\tt predict}$ and the $\text{\tt update}$ functions. The
overhead in the implementation and in the computational cost is minor.
Hence, the proposed method makes it straightforward to robustify any 
existing GF algorithm.

\section{Simulation Experiments}\label{sec:simulation}
In this section, we evaluate the RGF through simulations.
First, we show that the optimal feature enables a good fit of the approximate
belief to the exact posterior in the linear system used in previous sections. 
Secondly, we evaluate the sensitivity of the
RGF to the choice of tail parameters (Section \ref{sec:exp:parm}).
Finally, we show that the proposed feature \eqref{eq:optimal_feature}, which we designed for a linear system,
also allows for robustification in nonlinear systems (Section
\ref{sec:exp:nonlinear}).

We implemented Algorithm \ref{alg:rgf} using Monte Carlo as method for the
numeric integration required by the $\text{\tt predict}$ and the $\text{\tt
  update}$ functions\footnote{Code is available at \url{https://git-amd.tuebingen.mpg.de/amd-clmc/python_gaussian_filtering}}.

\subsection{Application to a Linear Filtering Problem}
\label{sec:exp:linear}

We revisit the simulation in Example \ref{system_specification} 
applying this time the RGF, using the true transition and sensor models. 
\begin{figure}[tb]
  \begin{subfigure}[b]{\linewidth}
    ~~
    \includegraphics[width=0.85\linewidth]{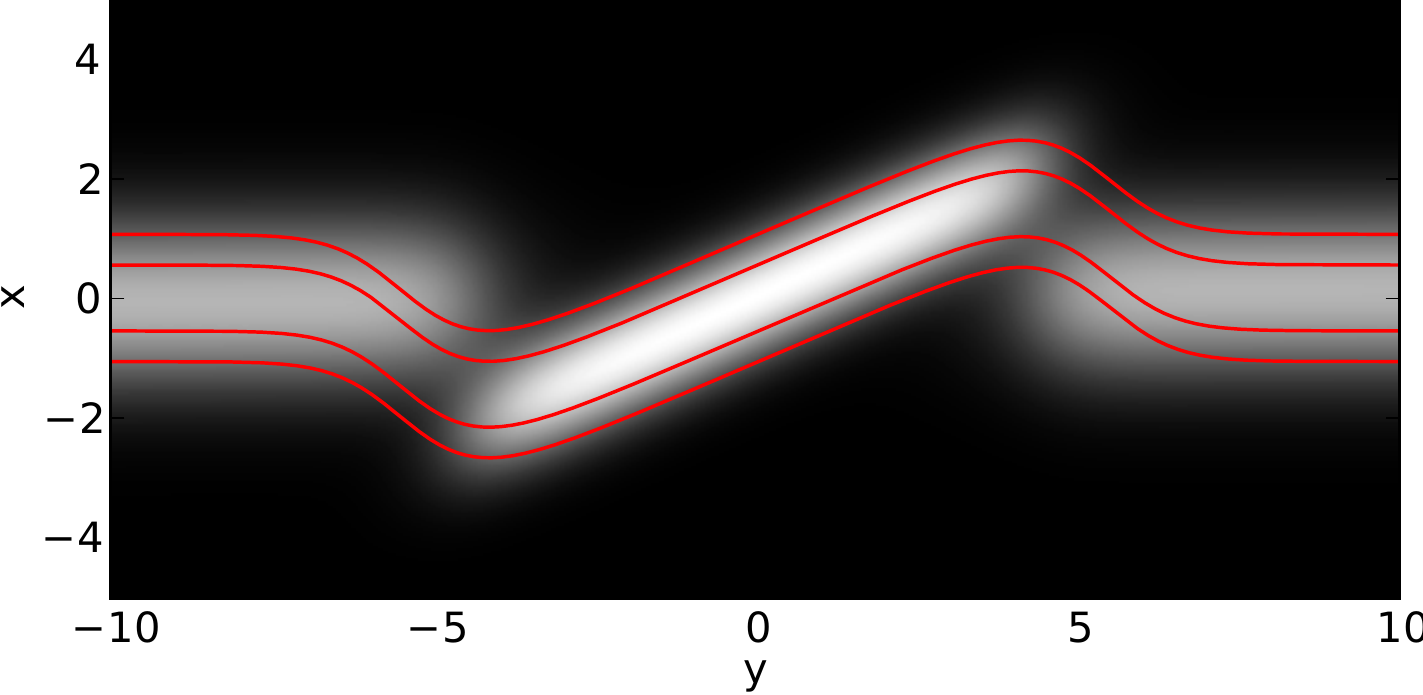}
	\caption{The contour lines of the approximate density $q(x_1|y_1)$
          overlaid on the exact density $p(x_1|y_1)$. \label{fig:rgf:density}}
  \end{subfigure}\\
  ~\\
  \begin{subfigure}[b]{\linewidth}
    ~~
    \includegraphics[width=0.85\linewidth]{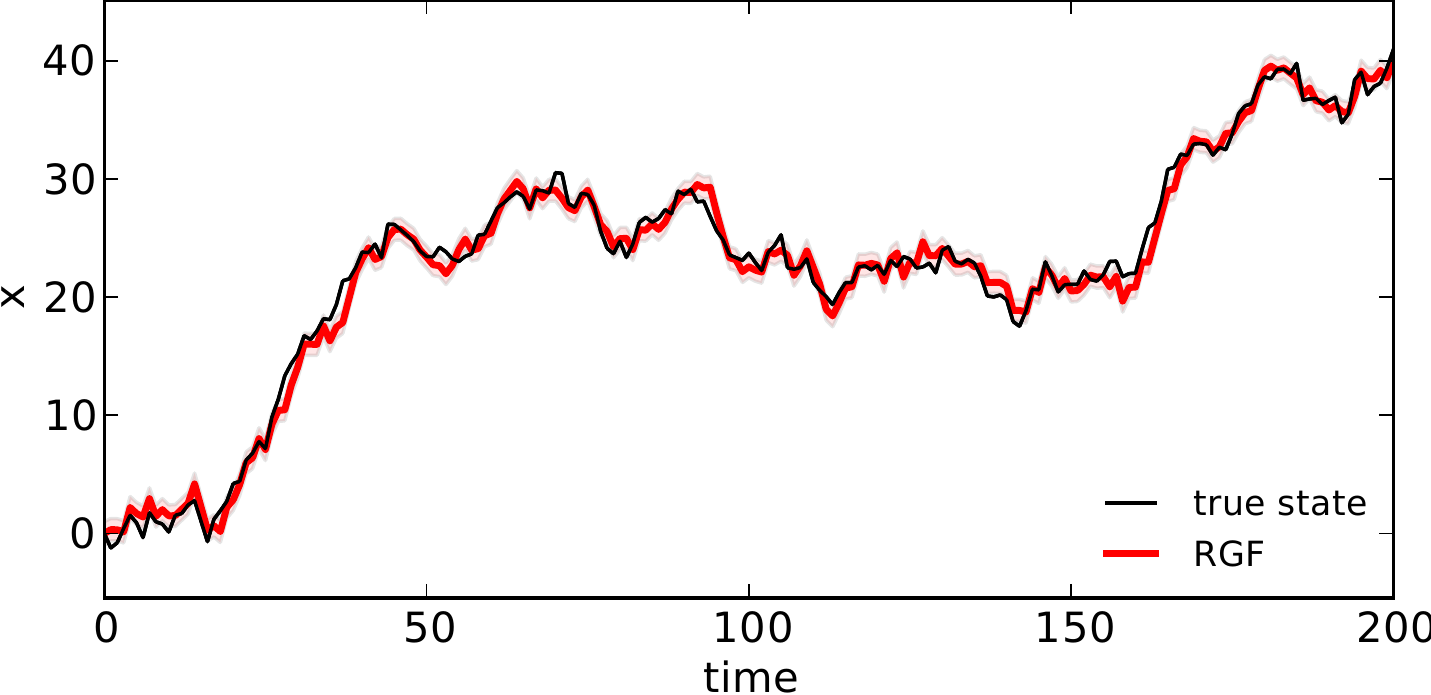}
	\caption{True state and filter estimates over time. \label{fig:rgf:time}}
  \end{subfigure}
  \caption{RGF applied to the system with fat-tailed measurement
    described in Example~\ref{system_specification}, 
    to be compared to the standard GF in Figure \ref{fig:gf_fat_tails}.}
\end{figure}
Comparing Figure \ref{fig:gf_fat_tails:density} to
Figure \ref{fig:rgf:density}, it is clear that the feature 
\eqref{eq:optimal_feature} allows for a much better fit of the approximation 
to the true density. As expected, this improved fit translates to a better 
filtering performance (Figure \ref{fig:rgf:time}). As desired, the proposed method is sensitive to 
measurements close to the expected values, but does not react to extreme values.

\subsection{Robustness to Tail Parameters}
\label{sec:exp:parm}
\begin{figure}[tb]
  \centering
  \includegraphics[width=0.85\linewidth]{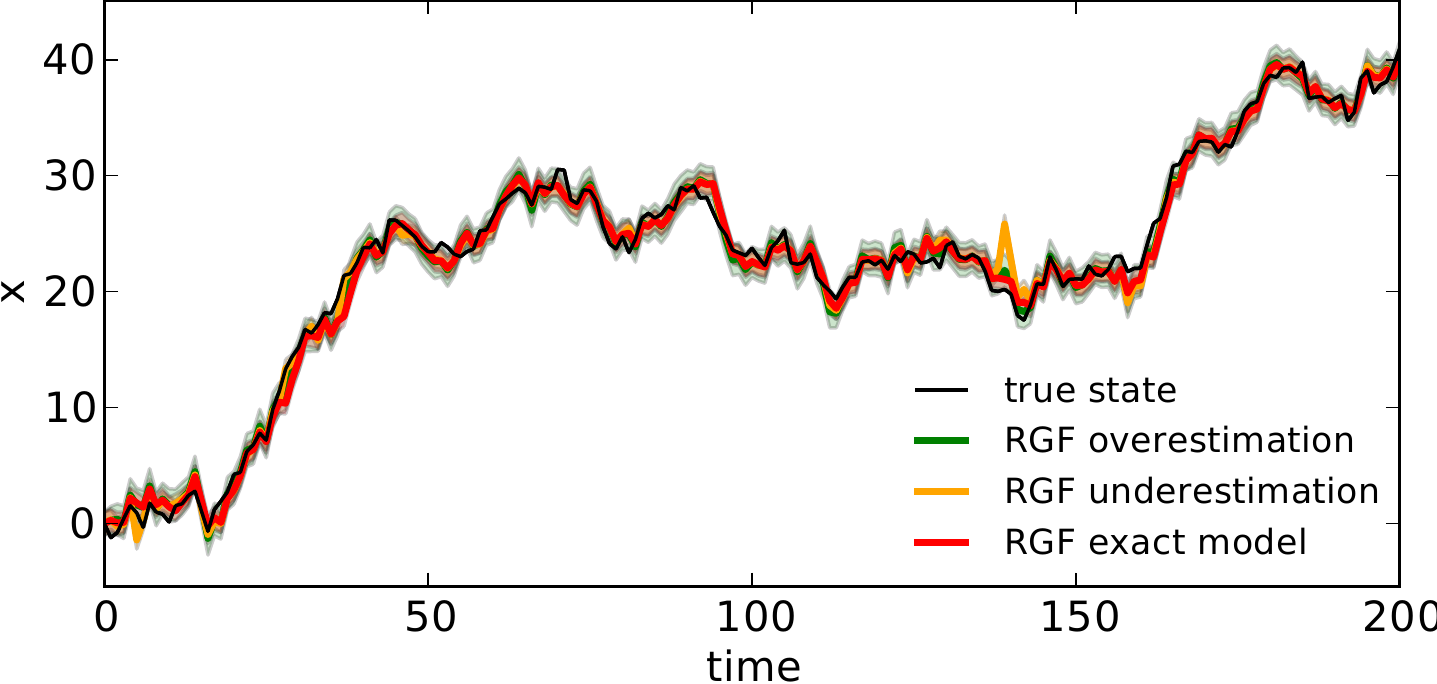}
  \caption{Robustness of the RGF to the choice of tail parameters. 
  The RGF
    behaves very similarly even when the tail parameters are severely under- or
    overestimated.
    \label{fig:parameter_robustness}}
\end{figure}
To show that the RGF is not sensitive to the specific choice of the tail
parameters, we simulate the same system as above, 
and run several RGFs with different tail parameters.
First, we apply a RGF
using a sensor model matching the true sensor, i.e. with tail
parameters $\omega = 0.1$, $\gamma = 10$. Then, we apply two RGFs which use incorrect
tail parameters. In one case we make both the weight and scale
of the tail much lower than in the true distribution: $\omega = 0.001$, $\gamma
= 1.0$ (underestimation of the true tail). In the other case we make them much higher: $\omega =
0.5$, $\gamma = 100.0$ (overestimation). Figure \ref{fig:parameter_robustness}
shows almost no degradation in the performance. 
The key aspect
enabling good filtering performance is that the sensor model has a tail
which decays slower than the Gaussian distribution, even when the shape of the
true tail is not precisely known.

\begin{table}
  \centering
  \scalebox{0.9}{
  \begin{tabular}{lrl|lrl}
    & value & units & 
    & value & units \\
    \hline
    $\Delta$                      &  $0.05$                  & s                            & 
    $\sigma_{\text{nom}, r}$      &  $0.5$                   & km                           \\
    $\sigma_v$                    &  $5 \cdot 10^{-3}$      & km/$\text{s}^2$              & 
    $\sigma_{\text{con}, r}$      &  $15.8$                  & km                           \\
    $\beta_0$                     &  $0.59$                  & $1$/km                       & 
    $\sigma_{\text{nom}, \theta}$  &  $0.63$                  & mrad                         \\
    $H_0$                         &  $13.4$                  & km                           & 
    $\sigma_{\text{con}, \theta}$ &  $200$                   & mrad                         \\
    $Gm_0$                        &  $3.986 \cdot 10^{5}$   & $\text{km}^{3}/\text{s}^{2}$ & 
    $\alpha$                      &  $0.15$                  &                              \\
    $R_0$                         &  $6374$                  & km                           & & & \\
    \end{tabular}
  }
  \caption{Simulation parameters. \label{tab:radar:constants}}
\end{table}
%
%
\begin{figure*}[tb]
  \centering
  \begin{subfigure}[b]{0.31\textwidth}
    \includegraphics[width=\textwidth]{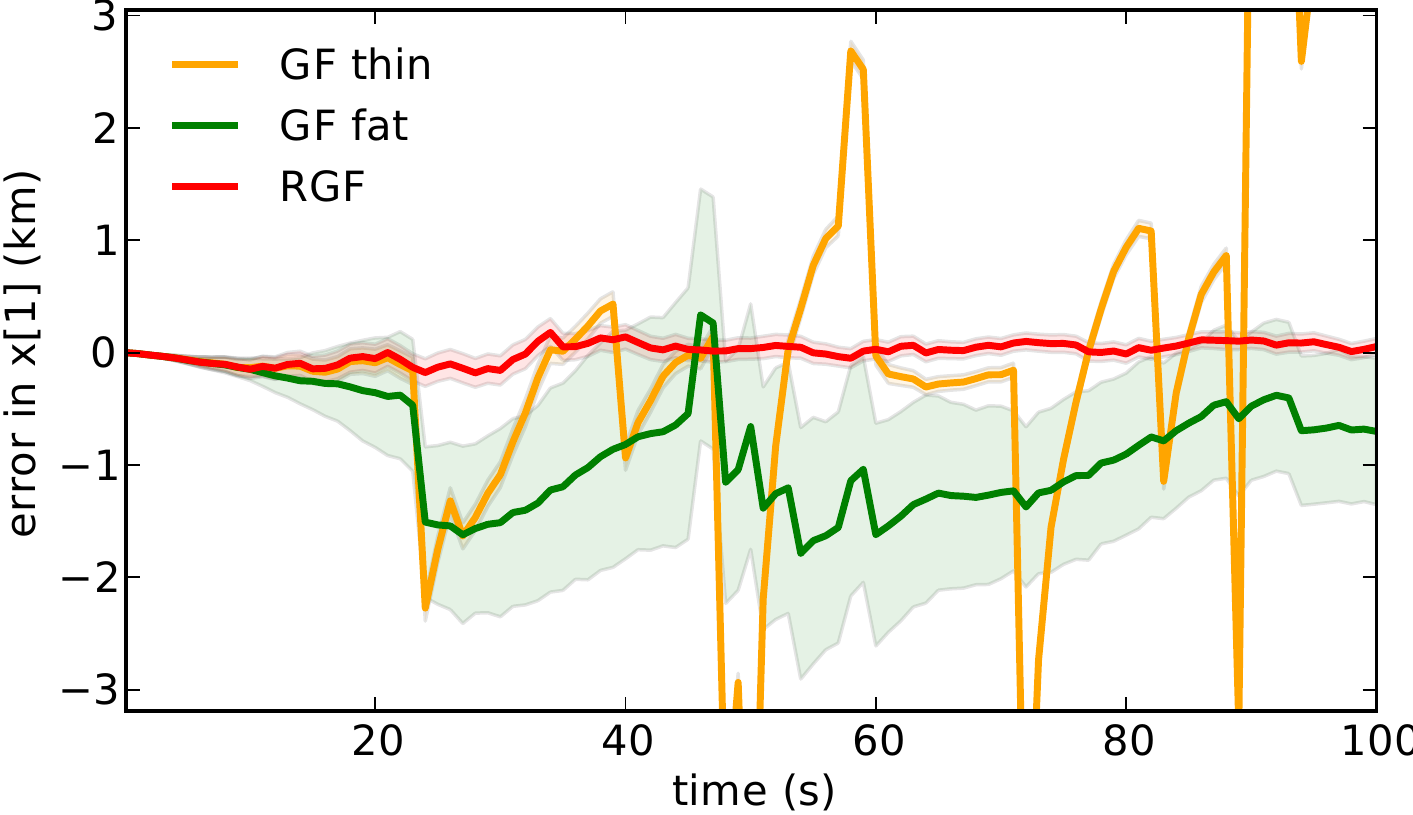}
	\caption{Error between the estimated and the true state $x^{[1]}$
          (position component).\label{fig:radar:x1}}
  \end{subfigure}
  ~
  \begin{subfigure}[b]{0.32\textwidth}
    \includegraphics[width=\textwidth]{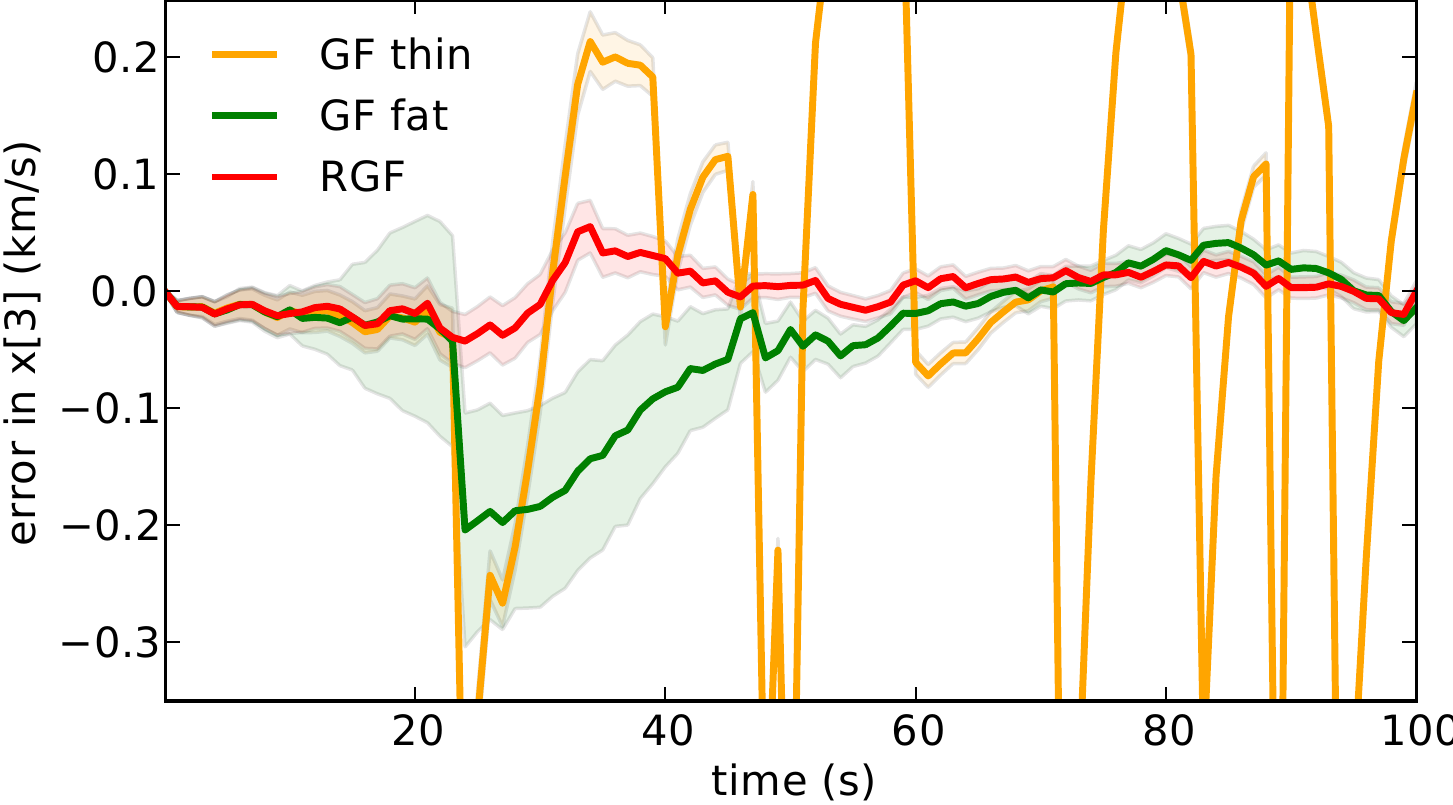}
	\caption{Error between estimated and true state $x^{[3]}$
          (velocity component).\label{fig:radar:x3}}
  \end{subfigure}
  ~
  \begin{subfigure}[b]{0.31\linewidth}
    \includegraphics[width=\linewidth]{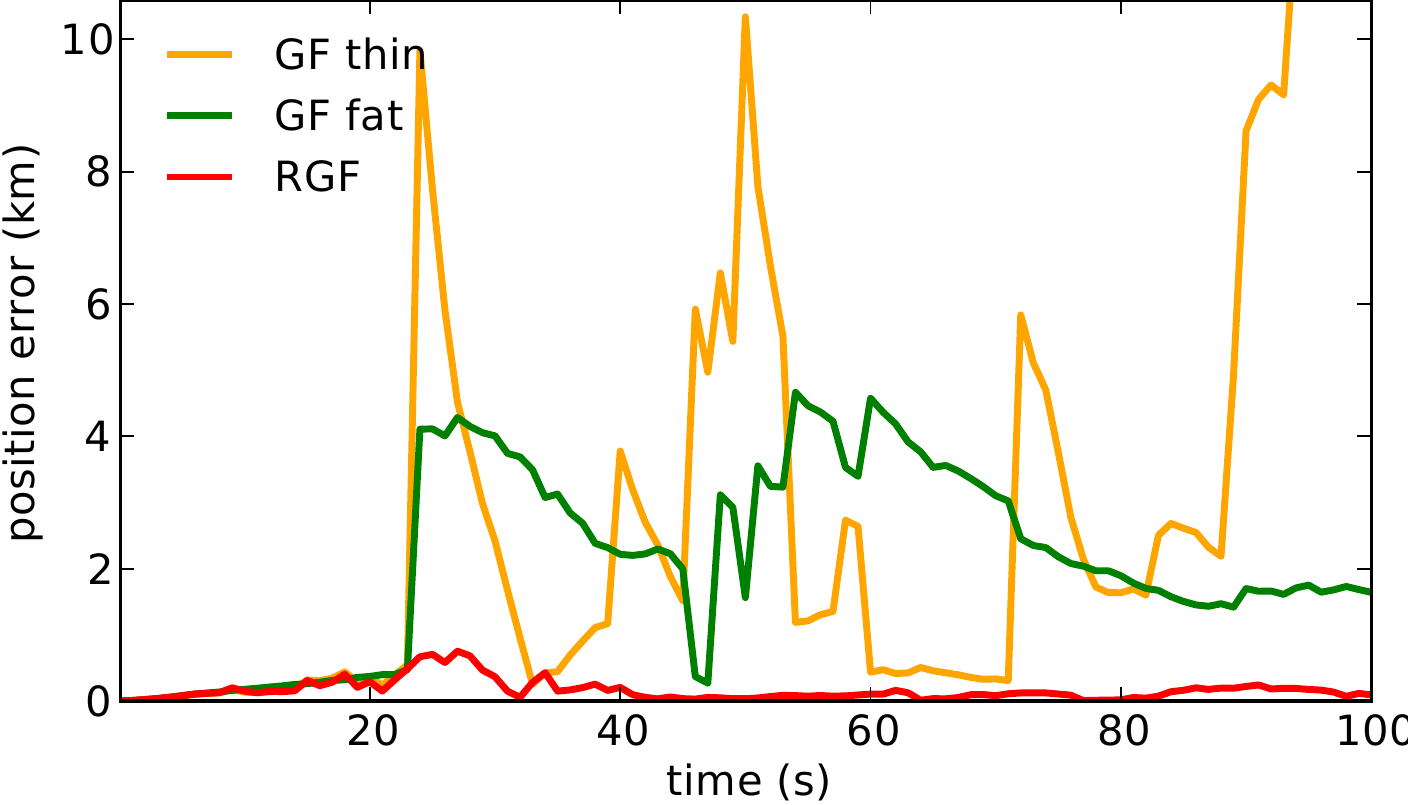}
	\caption{Euclidean distance between estimated and true 2D
          position.\label{fig:radar:position_err}}
  \end{subfigure}
  \caption{Results on the nonlinear filtering problem. The RGF deals well with the
    nonlinearities and the fat-tailed measurements.\label{fig:radar}}
\end{figure*}

\subsection{Application to a Nonlinear Filtering Problem}
\label{sec:exp:nonlinear}
As an example of nonlinear filtering, we consider 
the problem of using measurements from a radar ground
station to track the position of a vehicle that
enters the atmosphere at high altitude and speed. 
\cristina{removed: The main forces acting on the vehicle are the aerodynamic
drag and gravity.} 
The measurements
provided by the radar are range and bearing angle to the target
vehicle. This type of problem has been used before to compare the
capability of filters to deal with strong nonlinearities,
e.g. \cite{julier2004unscented,karlgaard}.

The noise in radar systems is typically referred to as glint noise in
the literature, and is known to be contaminated with outliers
\cite{hewer1987,wu1994,karlgaard,bilik2006,du2015}.  
It has been
modeled in different ways, e.g. using a Student $t$ distribution, or
as a mixture of two zero-mean Gaussian distributions (one with high
weight and low variance and another with low weight and high
variance), see \cite{bilik2006} and references therein.
%
%
%
In this section, we simulate the same system as in
\cite{julier2004unscented}, but replace their Gaussian measurement
noise with a mixture of two Gaussians as in
\cite{karlgaard,bilik2006}. 

%
\subsubsection*{State Transition Process}
The state consists of the position of the vehicle $(x^{[1]}, x^{[2]})$,
its velocity ($x^{[3]}$, $x^{[4]}$), and an unknown aerodynamics parameter $x^{[5]}$,
which has to be estimated. The state dynamics are
\begin{align}
  {x}^{[1]}_{t+1} &= {x}^{[1]}_{t} + \Delta x^{[3]}_t \label{eq:radar:dyn:0}\\
  {x}^{[2]}_{t+1} &= {x}^{[2]}_{t} + \Delta x^{[4]}_t \\
  {x}^{[3]}_{t+1} &= {x}^{[3]}_{t} + \Delta (D_t x^{[3]}_t + G_t x^{[1]}_t) +  \sqrt{\Delta}~\sigma_v~v^{[1]}_t \\
  {x}^{[4]}_{t+1} &= {x}^{[4]}_{t} + \Delta (D_t x^{[4]}_t + G_t x^{[2]}_t) + \sqrt{\Delta}~\sigma_v~v^{[2]}_t \\
  {x}^{[5]}_{t+1} &= {x}^{[5]}_{t},  \label{eq:radar:dyn:4}
\end{align}               
where $v^{[1]}$ and $v^{[2]}$ follow a standard normal distribution, and $\Delta$
is the discretization time step. 
The drag and gravity coefficients, 
$D_t = - \beta_t \exp \left(\frac{R_0 - R_t}{H_0}\right) V_t$ 
and $G_t = - \frac{Gm_0}{R^{3}_t}$, 
depend on the distance of the object to the centre of the 
Earth $R_t = \sqrt{(x^{[1]}_t)^{2} + (x^{[2]}_t)^{2}}$, its
speed $V_t = \sqrt{(x^{[3]}_t)^{2} + (x^{[4]}_t)^{2}}$, and 
its unknown ballistic coefficient $\beta_t = \beta_0 \exp( x^{[5]}_t)$.
Other quantities such
as the nominal ballistic coefficient $\beta_0$ and the mass $m_0$ and and radius
$R_0$ of the Earth are constant, see Table~\ref{tab:radar:constants}.  

\subsubsection*{Sensor Model}
The radar is located at $(x_r, y_r)$ and measures range $r_t$ and
bearing angle $\theta_t$ to the target object
\begin{align}
  r_t &= \sqrt{(x^{[1]}_t - x_r)^{2} + (x^{[2]}_t - y_r)^{2}} + w^{[1]}_t\\
  \theta_t &= 10^{3} \arctan \left(\frac{x^{[2]}_t - y_r}{x^{[1]}_t - x_r} \right) 
    + w^{[2]}_t \\
  w &\sim (1 - \alpha) \mathcal{N}(w | 0, \Sigma_{\text{nom}}) 
    + \alpha \mathcal{N}(w | 0, \Sigma_{\text{con}}) \label{eq:measurement_noise}. 
\end{align}
The nominal noise covariance is represented by 
$\Sigma_{\text{nom}} = \text{diag}([\sigma_{\text{nom}, r}^2, \sigma_{\text{nom}, \theta}^2])$, and
$\Sigma_{\text{con}} =  \text{diag}([\sigma_{\text{con}, r}^2, \sigma_{\text{con}, \theta}^2])$ 
is the covariance of the contaminating noise component. We
use $\alpha$ and covariances similar to \cite{bilik2006}, see Table
\ref{tab:radar:constants}.

\subsubsection*{Filter Specification}
We compare an RGF with two GFs. The three filters use transition models that
coincide with the real process \eqref{eq:radar:dyn:0}--\eqref{eq:radar:dyn:4}.

In problems of this type, the contaminating noise is often not precisely known. 
Therefore, we make our RGF assume a measurement
model as in Example \ref{system_specification}
\begin{align}
  &p_{\text{RGF}}(w) = 0.9~\mathcal{N}(w |0, \Sigma_{\text{nom}}) \nonumber \\ 
  & \;\;\;\; + 0.1~\mathcal{C}(w|0, \text{diag}([(10\sigma_{\text{nom}, r})^2, (10\sigma_{\text{nom}, \theta})^2])),
\end{align}
which makes use of some knowledge of the nominal noise $\Sigma_{\text{nom}}$,
while the shape of the tail and the mixing weight take default values.
Similarly, the first GF only knows about $\Sigma_{\text{nom}}$
\begin{align}
  p_{\text{GFthin}}(w) = \mathcal{N}(w |0, \Sigma_{\text{nom}}).
\end{align}

As discussed in Section \ref{sec:gf_fat_tails}, 
the GF is not able to produce accurate estimates in systems with large variance
even if the true measurement process \eqref{eq:measurement_noise} is known.
To show this empirically, we apply a second GF which uses the true covariance of the
sensor \eqref{eq:measurement_noise}
\begin{align}
   p_{\text{GFfat}}(w) = \mathcal{N}(w |0, (1 - \alpha) 
    \Sigma_{\text{nom}} + \alpha \Sigma_{\text{con}}).
\end{align}

We simulate the system during 100~s, using the integration time step $\Delta$
for the predictions and taking radar measurements at 1~Hz. 
As in \cite{julier2004unscented}, the initial state of
the system is $x_0 = [6500.4, 349.14, -1.8093, -6.7967, 0.6932]$, and the
initial belief for all filters is centered at $\mu_0 = [6500.4, 349.14,
  -1.8093, -6.7967, 0]$. Note the mismatch between the true ballistic
coefficient and the initial belief, i.e. the nominal $\beta_0$.

\subsubsection*{Results}
Figures \ref{fig:radar:x1} and \ref{fig:radar:x3} respectively show the error in the 
estimate of $x^{[1]}$ and the corresponding velocity $x^{[3]}$. We do not include the 
error in the position and velocity along the other dimension, since
they are qualitatively similar.
We can see that the GF using the
nominal variance (yellow) reacts strongly to outliers. The GF using the true
variance (green) of the sensor does not react as strongly. However,
due to the large variance, it tracks the true state poorly. In contrast, the RGF
(red) is robust to outliers and at the same time tracks the true state
well. This translates to a low 2D location error as shown in Figure
\ref{fig:radar:position_err}.
These results indicate that the optimal feature for linear systems
allows to robustify nonlinear systems too.
\section{Conclusion} \label{sect:discussion}
In the standard GF algorithm, the mean estimate is an affine function
of the measurement. We showed that for fat-tailed sensor models
this provides a very poor approximation to the exact posterior mean.

A recent result \cite{new_perspective} showed that filtering in measurement feature 
space can allow for more 
accurate approximations of the exact posterior.
Here, we have found the feature
that is optimal for fat-tailed sensor models under certain conditions.

We have shown both theoretically and in simulation that applying the standard
GF in this feature space enables it to work well with fat-tailed 
sensor models. The proposed RGF is hence robust to outliers while 
maintaining the computational efficiency of the standard GF. Any member of the 
family of GFs, such as the EKF or the UKF, can thus be robustified
by the proposed method without changing any of the main computations.

We have applied this algorithm to the problem of 3D object tracking
using an Xtion range sensor \cite{tracking_shizzle}. 
The main source of outliers in this
application are occlusions of the tracked object. While the standard GF
immediately loses track of the object when occlusions occur, the RGF
works well even under heavy occlusion.

%



\scriptsize{
  \bibliographystyle{unsrtnat}
\bibliography{bibliography}

\begin{thebibliography}{33}
\providecommand{\natexlab}[1]{#1}
\providecommand{\url}[1]{\texttt{#1}}
\expandafter\ifx\csname urlstyle\endcsname\relax
  \providecommand{\doi}[1]{doi: #1}\else
  \providecommand{\doi}{doi: \begingroup \urlstyle{rm}\Url}\fi

\bibitem[Sorenson(1960)]{ekf}
H.~W. Sorenson.
\newblock \emph{Kalman Filtering: Theory and Application}.
\newblock IEEE Press selected reprint series. IEEE Press, 1960.

\bibitem[Julier and Uhlmann(1997)]{ukf}
S.~J. Julier and J.~K. Uhlmann.
\newblock A new extension of the {K}alman filter to nonlinear systems.
\newblock In \emph{Proceedings of AeroSense: The 11th Int. Symp. on
  Aerospace/Defense Sensing, Simulations and Controls}, pages 182--193, 1997.

\bibitem[S{\"a}rkk{\"a}(2013)]{sarkka}
S.~S{\"a}rkk{\"a}.
\newblock \emph{Bayesian filtering and smoothing}.
\newblock Cambridge University Press, New York, NY, USA, 2013.

\bibitem[Schick and Mitter(1994)]{schick}
I.~C. Schick and S.~K. Mitter.
\newblock Robust recursive estimation in the presence of heavy-tailed
  observation noise.
\newblock \emph{The Annals of Statistics}, 1994.

\bibitem[Huber(1964)]{huber1964}
P.~J. Huber.
\newblock Robust estimation of a location parameter.
\newblock \emph{Annals of Mathematical Statistics}, 1964.

\bibitem[Masreliez and Martin(1977)]{masreliez}
C.~Masreliez and R.~Martin.
\newblock Robust {B}ayesian estimation for the linear model and robustifying
  the {K}alman filter.
\newblock \emph{IEEE Transactions on Automatic Control}, 1977.

\bibitem[Roth et~al.(2013)Roth, Ozkan, and Gustafsson]{roth}
M.~Roth, E.~Ozkan, and F.~Gustafsson.
\newblock A {S}tudent's t filter for heavy tailed process and measurement
  noise.
\newblock In \emph{IEEE International Conference on Acoustics, Speech and
  Signal Processing (ICASSP)}, 2013.

\bibitem[Meinhold and Singpurwalla(1989)]{meinhold}
R.~J. Meinhold and N.~D. Singpurwalla.
\newblock Robustification of {K}alman filter models.
\newblock \emph{Journal of the American Statistical Association}, 1989.

\bibitem[Ting et~al.(2007)Ting, Theodorou, and Schaal]{ting}
J.-A. Ting, E.~Theodorou, and S.~Schaal.
\newblock A {K}alman filter for robust outlier detection.
\newblock In \emph{IEEE/RSJ International Conference on Intelligent Robots and
  Systems (IROS)}, 2007.

\bibitem[S{\"a}rkk{\"a} and Nummenmaa(2009)]{sarkka_noise}
S.~S{\"a}rkk{\"a} and A.~Nummenmaa.
\newblock Recursive noise adaptive {K}alman filtering by variational {B}ayesian
  approximations.
\newblock \emph{IEEE Transactions on Automatic Control}, 2009.

\bibitem[Agamennoni et~al.(2011)Agamennoni, Nieto, and Nebot]{agamennonizzle}
G~Agamennoni, J.~I. Nieto, and E.~M. Nebot.
\newblock An outlier-robust kalman filter.
\newblock In \emph{Robotics and Automation (ICRA), 2011 IEEE International
  Conference on}, 2011.

\bibitem[Agamennoni et~al.(2012)Agamennoni, Nieto, and Nebot]{agamennoni}
G~Agamennoni, J.~I. Nieto, and E.~M. Nebot.
\newblock Approximate inference in state-space models with heavy-tailed noise.
\newblock \emph{IEEE Transactions on Signal Processing}, 2012.

\bibitem[van~der Merwe and Wan(2003)]{spkf}
R.~van~der Merwe and E.~Wan.
\newblock {Sigma-Point Kalman Filters} for probabilistic inference in dynamic
  state-space models.
\newblock In \emph{In Proceedings of the Workshop on Advances in Machine
  Learning}, 2003.

\bibitem[Ito and Xiong(2000)]{gf}
K.~Ito and Kaiqi Xiong.
\newblock {Gaussian filters for nonlinear filtering problems}.
\newblock \emph{{IEEE Transactions on Automatic Control}}, 45\penalty0
  (5):\penalty0 910--927, May 2000.

\bibitem[N{\o}Rgaard et~al.(2000)N{\o}Rgaard, Poulsen, and Ravn]{ddf}
Magnus N{\o}Rgaard, Niels~K. Poulsen, and Ole Ravn.
\newblock New developments in state estimation for nonlinear systems.
\newblock \emph{Automatica}, 36\penalty0 (11):\penalty0 1627--1638, November
  2000.
\newblock ISSN 0005-1098.

\bibitem[Karlgaard and Schaub(2006)]{karlgaard}
C.~D. Karlgaard and H.~Schaub.
\newblock Comparison of several nonlinear filters for a benchmark tracking
  problem.
\newblock In \emph{AIAA Guidance, Navigation, and Control Conference and
  Exhibit}, Keystone, CO, USA, august 2006.

\bibitem[Piche et~al.(2012)Piche, S{\"a}rkk{\"a}, and Hartikainen]{piche}
R.~Piche, S.~S{\"a}rkk{\"a}, and J.~Hartikainen.
\newblock Recursive outlier-robust filtering and smoothing for nonlinear
  systems using the multivariate student-t distribution.
\newblock In \emph{IEEE International Workshop on Machine Learning for Signal
  Processing (MLSP)}, 2012.

\bibitem[Kushner(1967)]{earlyKushner}
H.~J. Kushner.
\newblock Approximations to optimal nonlinear filters.
\newblock \emph{IEEE Transactions on Automatic Control}, 12\penalty0
  (5):\penalty0 546--556, 1967.

\bibitem[Kalman(1960)]{kalman1960new}
R.~E. Kalman.
\newblock {A New Approach to Linear Filtering and Prediction Problems}.
\newblock \emph{Transactions of the ASME - Journal of Basic Engineering},
  \penalty0 (82 (Series D)):\penalty0 35--45, 1960.

\bibitem[W\"uthrich et~al.(2015)W\"uthrich, Trimpe, Kappler, and
  Schaal]{new_perspective}
M.~W\"uthrich, S.~Trimpe, D.~Kappler, and S.~Schaal.
\newblock {A New Perspective and Extension of the Gaussian Filter}.
\newblock In \emph{{Robotics: Science and Systems (R:SS)}}, 2015.

\bibitem[Wu et~al.(2006)Wu, Hu, Wu, and Hu]{wu}
Y.~Wu, D.~Hu, M.~Wu, and X.~Hu.
\newblock A numerical-integration perspective on {G}aussian filters.
\newblock \emph{IEEE Transactions on Signal Processing}, 54\penalty0
  (8):\penalty0 2910--2921, 2006.

\bibitem[MacKay(2003)]{mckay}
David~J. MacKay.
\newblock \emph{Information Theory, Inference and Learning Algorithms}.
\newblock Cambridge University Press, 2003.

\bibitem[Arasaratnam and Haykin(2009)]{ckf}
I.~Arasaratnam and S.~Haykin.
\newblock Cubature {K}alman filters.
\newblock \emph{Automatic Control, IEEE Transactions on}, 2009.

\bibitem[de~Finetti(1937)]{definetti}
B.~de~Finetti.
\newblock La pr{\'e}vision~: ses lois logiques, ses sources subjectives.
\newblock \emph{Annales de l'institut {H}enri Poincar{\'e}}, 1937.

\bibitem[Cooke et~al.(2011)Cooke, Nieboer, and Misiewicz]{fat_tails}
R.~Cooke, D.~Nieboer, and J.~Misiewicz.
\newblock Fat-tailed distributions: Data, diagnostics, and dependence.
\newblock Technical report, 2011.

\bibitem[Barber(2012)]{barber}
D.~Barber.
\newblock \emph{{Bayesian Reasoning and Machine Learning}}.
\newblock {Cambridge University Press}, New York, NY, USA, 2012.

\bibitem[Thrun et~al.(2001)Thrun, Fox, Burgard, and
  Dellaert]{thrun_monte_carlo}
S.~Thrun, D.~Fox, W.~Burgard, and F.~Dellaert.
\newblock Robust {Monte Carlo} localization for mobile robots.
\newblock \emph{Artificial Intelligence}, 2001.

\bibitem[Julier and Uhlmann(2004)]{julier2004unscented}
S.~J. Julier and J.~K. Uhlmann.
\newblock Unscented filtering and nonlinear estimation.
\newblock \emph{Proceedings of the IEEE}, 2004.

\bibitem[Hewer et~al.(1987)Hewer, Martin, and Zeh]{hewer1987}
G.~A. Hewer, R.~D. Martin, and J.~Zeh.
\newblock Robust preprocessing for {K}alman filtering of glint noise.
\newblock \emph{IEEE Transactions on Aerospace and Electronic Systems}, 1987.

\bibitem[Wu and Cheng(1994)]{wu1994}
W.-R. Wu and P.-P. Cheng.
\newblock A nonlinear {IMM} algorithm for maneuvering target tracking.
\newblock \emph{IEEE Transactions on Aerospace and Electronic Systems}, 1994.

\bibitem[Bilik and Tabrikian(2006)]{bilik2006}
I.~Bilik and J.~Tabrikian.
\newblock Target tracking in glint noise environment using nonlinear
  non-{G}aussian {K}alman filter.
\newblock In \emph{IEEE Conference on Radar}, 2006.

\bibitem[Du et~al.()Du, Wang, and Bai]{du2015}
H.~Du, W.~Wang, and L.~Bai.
\newblock Observation noise modeling based particle filter: An efficient
  algorithm for target tracking in glint noise environment.
\newblock \emph{Neurocomputing}.

\bibitem[Issac et~al.(2016)Issac, W\"uthrich, Garcia~Cifuentes, Bohg, Trimpe,
  and Schaal]{tracking_shizzle}
J.~Issac, M.~W\"uthrich, C.~Garcia~Cifuentes, J.~Bohg, S.~Trimpe, and
  S.~Schaal.
\newblock {Depth-Based Object Tracking Using a Robust Gaussian Filter}.
\newblock In \emph{Robotics and Automation (ICRA), IEEE International
  Conference on}, 2016.
\newblock URL \url{http://arxiv.org/abs/1602.06157}.

\end{thebibliography}
}


%

\end{document}